\documentclass[letterpaper]{article} 
\usepackage{aaai2026}  
\usepackage{times}  
\usepackage{helvet}  
\usepackage{courier}  
\usepackage[hyphens]{url}  
\usepackage{graphicx} 
\usepackage{booktabs}
\usepackage{multirow}
\usepackage{array}
\usepackage{natbib}  
\usepackage{caption} 
\usepackage{algorithm}
\usepackage{algorithmic}

\usepackage{newfloat}
\usepackage{listings}
\usepackage{amsmath}
\usepackage{amsfonts}
\DeclareCaptionStyle{ruled}{labelfont=normalfont,labelsep=colon,strut=off} 
\floatstyle{ruled}
\newfloat{listing}{tb}{lst}{}
\floatname{listing}{Listing}
\title{A HyperGraphMamba-Based Multichannel Adaptive Model for ncRNA Classification}
\author {
    Xin An\textsuperscript{\rm 1}\equalcontrib,
    Ruijie Li\textsuperscript{\rm 2}\equalcontrib,
    Qiao Ning\textsuperscript{\rm 3}\thanks{Corresponding author.},
    Hui Li\textsuperscript{\rm 1},
    Qian Ma\textsuperscript{\rm 1},
    Shikai Guo\textsuperscript{\rm 1}
}
\affiliations {
    
    \textsuperscript{\rm 1} Dalian Maritime University\\
    Dalian, China\\
    \textsuperscript{\rm 2} The Hong Kong University of Science and Technology (Guangzhou)\\
    Guangzhou, China\\
    \textsuperscript{\rm 3} Jiangnan University\\
    Wuxi, China\\
    \{anxin2695, li\_hui, maqian, shikai.guo\}@dlmu.edu.cn, rli541@connect.hkust-gz.edu.cn,
    ningq669@jiangnan.edu.cn
}
\usepackage{bibentry}

\begin{document}

\maketitle

\begin{abstract}
Non-coding RNAs (ncRNAs) play pivotal roles in gene expression regulation and the pathogenesis of various diseases. Accurate classification of ncRNAs is essential for functional annotation and disease diagnosis. To address existing limitations in feature extraction depth and multimodal fusion, we propose HGMamba-ncRNA, a HyperGraphMamba-based multichannel adaptive model, which integrates sequence, secondary structure, and optionally available expression features of ncRNAs to enhance classification performance. Specifically, the sequence of ncRNA is modeled using a parallel Multi-scale Convolution and LSTM architecture (MKC-L) to capture both local patterns and long-range dependencies of nucleotides. The structure modality employs a multi-scale graph transformer (MSGraphTransformer) to represent the multi-level topological characteristics of ncRNA secondary structures. The expression modality utilizes a Chebyshev Polynomial-based Kolmogorov–Arnold Network (CPKAN) to effectively model and interpret high-dimensional expression profiles. Finally, by incorporating virtual nodes to facilitate efficient and comprehensive multimodal interaction, HyperGraphMamba is proposed to adaptively align and integrate multichannel heterogeneous modality features. Experiments conducted on three public datasets demonstrate that HGMamba-ncRNA consistently outperforms state-of-the-art methods in terms of accuracy and other metrics. Extensive empirical studies further confirm the model’s robustness, effectiveness, and strong transferability, offering a novel and reliable strategy for complex ncRNA functional classification. Code and datasets are available at https://anonymous.4open.science/r/HGMamba-ncRNA-94D0.
\end{abstract}


\section{Introduction}

Non-coding RNAs (ncRNAs) were once considered non-functional elements of the genome~\cite{chen2022computational,alexander2010annotating}. However, in the 2000s, projects such as the Human Genome Project highlighted that approximately 98\% of the human genome is non-coding, thereby shifting scientific attention toward ncRNAs~\cite{mattick2001non}. Accumulating evidence has demonstrated that ncRNAs play critical roles in a wide range of biological processes and gene regulatory mechanisms. These include the regulation of physiological functions, development, and disease progression~\cite{beermann2016non}, as well as the promotion or suppression of various types of cancer~\cite{romano2017small}. Consequently, functional characterization of ncRNAs may contribute to the identification of novel biomarkers and therapeutic targets~\cite{li2025mhmda}. In computational biology, functional classification of ncRNAs facilitates large-scale role annotation. Such groups include families, defined by shared evolutionary origin, and broader categories, which group functionally similar ncRNAs and may span multiple families~\cite{creux2024comparison}.

\begin{figure}
\centering
\includegraphics[width=0.48\textwidth]{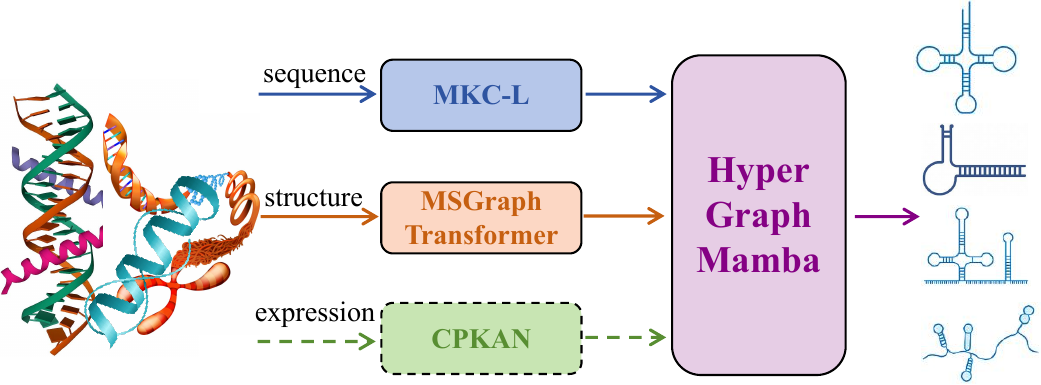}
\captionsetup{font=scriptsize}
\caption{A brief overview of the HGMamba-ncRNA framework.}
\label{fig:1}
\end{figure}
NcRNAs are broadly classified by transcript length into long non-coding RNAs (lncRNAs; $>200$ nt) and short non-coding RNAs ($<200$ nt)~\cite{shi2021genome,chen2019small}. LncRNAs primarily regulate gene expression via interactions with DNA, RNA, or proteins~\cite{wang2019review,bridges2021lnccation}, while short ncRNAs, such as microRNAs (miRNAs), typically act at the post-transcriptional level. Functionally, ncRNAs are categorized as housekeeping (e.g., rRNAs, tRNAs), essential for basic cellular activities, or regulatory (e.g., miRNAs, siRNAs), which modulate gene expression at multiple levels.  Despite progress in understanding certain ncRNAs, the functions of most remain unclear. High-throughput sequencing has identified numerous novel ncRNAs, posing challenges for their functional characterization and classification~\cite{guan2022non,panni2020non,vilaca2023strategies,xu2022opportunities}.

Various methods have been developed to classify ncRNAs into functional categories. Early approaches relied on probabilistic models, dynamic programming~\cite{lindgreen2007mastr}, or traditional machine learning (ML) algorithms~\cite{min2017deep}. ML techniques can effectively capture distinguishing features of ncRNAs to differentiate them from protein-coding sequences. More recently, deep learning (DL), a subfield of ML, has achieved notable success in this domain~\cite{baek2018lncrnanet}, with many subsequent methods adopting architectures~\cite{creux2024comparison,liu2022prediction,li2022lncdc} such as convolutional neural networks (CNNs) and recurrent neural networks (RNNs)~\cite{chaabane2020circdeep,liu2019prediction}.

Recent research increasingly emphasizes feature learning for ncRNA classification. Several methods focus on sequence-based features. For instance, RiNALMo~\cite{penic2025rinalmo} uses a BERT-style Transformer pre-trained on ncRNA corpora. ncrna-deep~\cite{noviello2020deep} and ncRDeep~\cite{chantsalnyam2020ncrdeep} utilize CNNs to directly extract features from raw sequences. BioDeepFuse~\cite{avila2024biodeepfuse} fuses handcrafted sequence features with CNN, BiLSTM, and attention. MFPred~\cite{chen2023mfpred} combines contextual and local sequence encodings. LMFE~\cite{zhang2025lmfe} fuses global and local features through multi-scale convolution and fully connected layers. Compared with sequences, structures contain more functional information. Therefore, some approaches leverage secondary structure for feature learning. RNAGCN~\cite{deng2022rnagcn} models dot-bracket structures as graphs and applies GCNs for feature extraction. Furthermore, some methods integrate both sequence and structure features~\cite{wu2016non}. MM-ncRNAFP~\cite{xu2024improving} combines sequence, structure, and pre-trained embeddings with attention-based fusion. MMnc~\cite{creux2025mmnc} further incorporates expression data in a multi-modal framework to comprehensively represent ncRNA characteristics.

Despite progress in deep learning for RNA classification, many methods underexploit diverse features and their interdependencies. To address this, we propose HGMamba-ncRNA (Figure~\ref{fig:1}). It uses a Multi-Kernel CNN and LSTM to extract local and global sequence patterns, a multi-scale graph Transformer for structural features, and CPKAN for expression data—adaptively modeling global trends and fine-grained variation. Finally, a HyperGraphMamba module fuses modalities to capture intrinsic relationships and boost classification. The key contributions are as follows:

\begin{itemize}
    \item We propose a Chebyshev Polynomial-based Kolmogorov–Arnold Network (CPKAN) that improves modeling capacity and robustness of high-dimensional RNA expression data while maintaining interpretability.
    \item We construct a Multi-Scale Graph Topological Transformer module (MSGraphTransformer) that incorporates learnable hop-based weights and edge-attribute attention mechanisms, enabling precise capture of both local and global topological features of RNA secondary structures.
    \item We design a parallel architecture, MKC-L, integrating Multi-Scale CNN and LSTM to jointly model multi-scale local features and global dependencies in RNA sequences, thereby improving semantic representation.
    \item We introduce a novel HyperGraphMamba model based on Mamba and hypergraph fusion, which—for the first time—incorporates hypernodes to achieve efficient alignment and deep interaction of multi-modal features.
\end{itemize}




\section{Methodology}

\begin{figure*}
\centering
\includegraphics[width=0.8\textwidth]{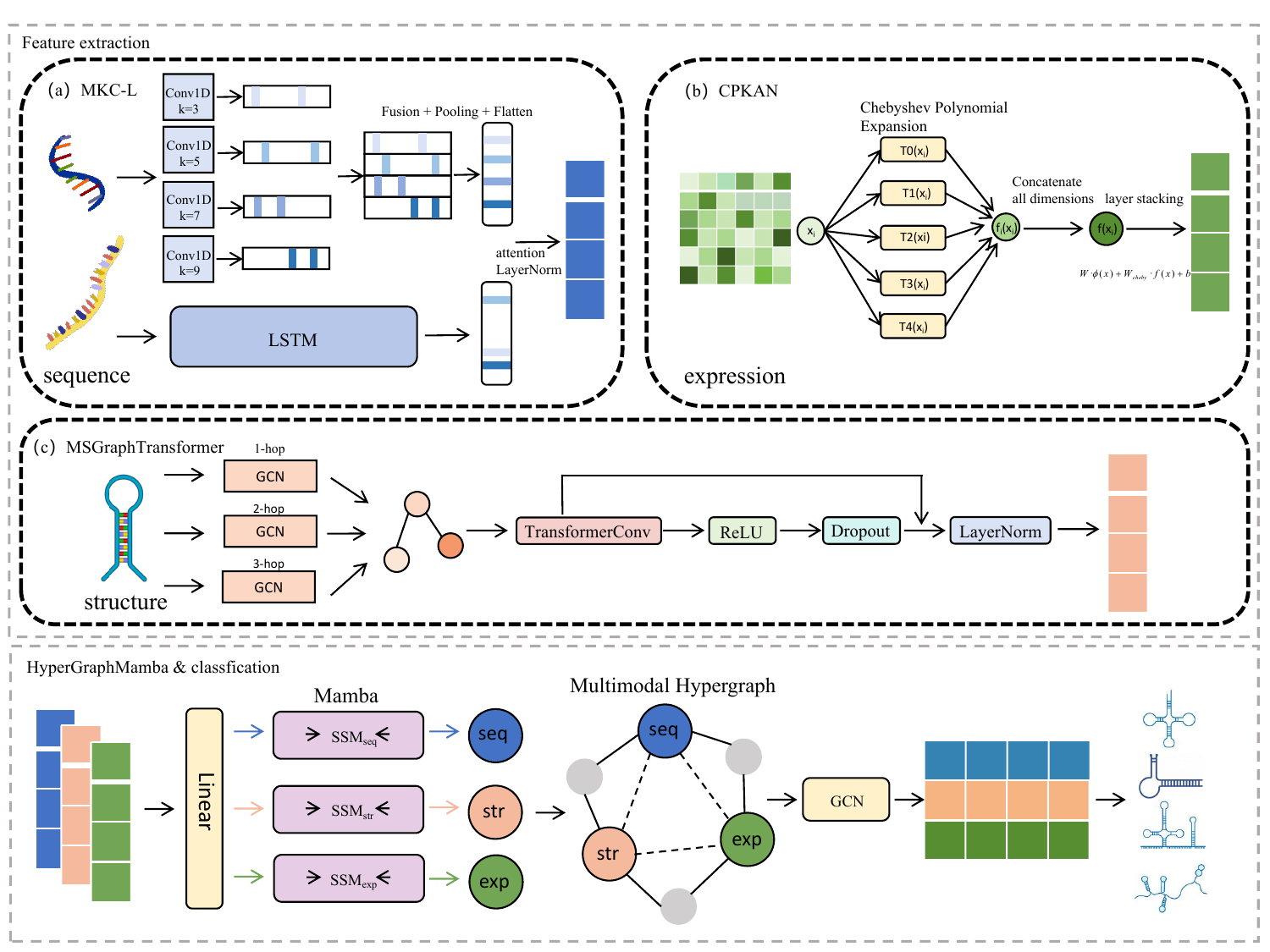}
\captionsetup{font=scriptsize}
\caption{Workflow of HGMamba-ncRNA.}
\label{fig:2}
\end{figure*}

The framework consists of three main components: 1) Multimodal Feature Extraction; 2) HyperGraphMamba Multimodal Feature Fusion; 3) ncRNA Classification. The structure of the model is shown in Figure~\ref{fig:2}.

\subsection{Multimodal Feature Extraction}

\subsubsection{CPKAN Module for Expression Features}

We propose CPKAN, a Chebyshev Polynomials-driven Kolmogorov–Arnold Network that replaces B-splines in classical KAN~\cite{liu2024kan} with Chebyshev orthogonal polynomials. Exploiting their optimal approximation and recursive orthogonality on [-1,1], CPKAN improves modeling of high-dimensional biological data while retaining interpretability. Tanh normalization reduces noise and outlier effects, and low-order expansions align with the sparsity of gene regulatory interactions. The orthogonal basis supports multi-scale, interpretable modeling from basic transcription to high-order regulation. To capture local semantic dependencies like co-expression and regulation, CPKAN learns independent nonlinear projections for each input dimension. 
We adopt Kaiming initialization for base weights and initialize Chebyshev coefficients using a normal distribution with mean 0 and standard deviation $\text{scale}_{\text{cheby}}/\sqrt{d}$.

Specifically, for the obtained representation $x$ of sample $X$, each dimension $x_i$ is first mapped to the standard interval [-1,1], and then expanded into a high-order representation using first-kind Chebyshev polynomials, defined as $T_n(x) = \cos(n \arccos(x))$. The expansion includes terms from
$T_0(x_i)$ to $T_N(x_i)$, forming a vector of Chebyshev basis functions for each input dimension.

A set of learnable coefficients $\alpha_{i,n}$ constructs a local nonlinear mapping $f_i(x_i)$ as a weighted sum over the Chebyshev terms. The full expanded representation $f(x)$ is then obtained by concatenating the outputs across all dimensions.

To integrate this with the base network, each layer fuses the activation output $\phi(x)$ and the Chebyshev-expanded features via a linear transformation:

\begin{equation}
h^{(l)} = W \cdot \phi(x) + W_{\text{cheby}} \cdot f(x) + b
\end{equation}
where $\phi$ denotes the base activation function. By stacking multiple such layers, the network performs deep nonlinear modeling from raw input to final embedding. The last-layer output $h^{(L)}$ serves as the expression modality feature:
  
\begin{equation}
F_{\text{exp}} = h^{(L)} 
\end{equation}


\subsubsection{MSGraphTransformer Module for Structural Features}

We propose MSGraphTransformer, a Multi-scale Graph Topology Transformer that captures multi-hop interactions and edge features in RNA secondary structures. A learnable scale-weight mechanism adaptively balances different-hop neighborhoods, while a multi-scale GCN encoder extracts hierarchical features from local base-pairing to global topology. An edge-aware TransformerConv~\cite{vaswani2017attention} integrates self-attention, preserving GCN smoothing and modeling long-range interactions. This design effectively captures local-global structural patterns, boosting representation and generalization in RNA structure embedding.

Specifically, first, multi-scale molecular graph adjacency structures $A^{(s)}$ are constructed based on different hop counts $s \in \{1,\, 2,\, 3\}$. For each scale graph, the GCNConv module is used to extract multi-scale node representations:
\begin{equation}
H^{(s)} = \sigma\left( \hat{D}^{(s)\,-\frac{1}{2}} \hat{A}^{(s)} \hat{D}^{(s)\,-\frac{1}{2}} X W^{(s)} \right)
\end{equation}
where $\hat{A}^{(s)}=A^{(s)}+1$ is the adjacency matrix with self-loops added, $\hat{D}^{(s)}$ is the corresponding degree matrix, $W^{(s)}$  is the learnable transformation matrix for each scale, and $\sigma(x)$ is the activation function.

The extracted multi-scale features are fused through learnable scale weights $\omega$:
\begin{equation}
H_{\text{fused}} = \sum_{s=1}^{|s|} \omega_s H^{(s)}
\end{equation}
The fused features are processed by a multi-layer TransformerConv module to model long-range dependencies between bases and the influence of edge attributes. The attention weight calculation for any pair of connected bases $i,j$ in the Transformer layer is as follows:
\begin{equation}
\alpha_{i,j} = \operatorname{softmax}_j \left( \frac{(W_Q h_i)^T (W_K h_j + W_E e_{ij})}{\sqrt{d}} \right)
\end{equation}
where $W_Q$, $W_K$ and $W_E$ are projection matrices for queries, keys, and edge attributes $e_{ij}$; $d$ is the attention dimension; and $h_i$ is the hidden state of base $i$.

The updated feature representation of base $i$ is:
\begin{equation}
h'_i = \sum_{j \in \mathcal{N}(i)} \alpha_{ij} \cdot (W_V h_j)
\end{equation}
where $W_V$ is the value transformation matrix.

The output of each Transformer layer undergoes residual connection and LayerNorm normalization:
\begin{equation}
H^{(l)} = \operatorname{LayerNorm}\left(H^{(l-1)} + \Delta H^{(l)}\right)
\end{equation}
where $\Delta H^{(l)}$ denotes the TransformerConv output with dropout. After stacking $L$ such layers, we obtain $H^{(L)}$, which serves as the structural modality feature $F_{str}$.

\subsubsection{MKC-L Module for Sequence Features}

We propose MKC-L, a "Multi-scale Convolution + LSTM" parallel fusion framework to capture local and global semantics of RNA sequences. The MultiScaleCNN module uses parallel branches with four kernel sizes (3, 5, 7, 9) and an inverse kernel size-based dynamic channel allocation strategy, enabling adaptive extraction of features from single-base modifications to long-range structures. Meanwhile, the LSTM module~\cite{hochreiter1997long} models global sequence dependencies, compensating for CNN’s limitations in long-range context. The two branches are fused via linear mapping and LayerNorm, preserving local sensitivity and global coherence, enhancing RNA sequence representation and downstream task performance.

The MultiScaleCNN module applies multiple 1D convolution kernels with different sizes (e.g., $k$=3,5,7,9) in parallel to extract multi-scale sequence features. Each convolution branch applies activation and batch normalization, and the resulting features are concatenated along the channel dimension. A subsequent 1×1 convolution is used for feature compression. To balance feature contributions across different kernel sizes, the number of output channels $c_i$ for each branch is inversely proportional to $k_i + 1$, normalized by a factor $Z = \sum_{j=1}^{4} \frac{1}{k_j + 1}$, ensuring the total number of output channels equals $C_{out}$.

To capture long-range dependencies, the original sequence $X \in \mathbb{R}^{B \times L \times C}$, where $B$ is the batch size, $L$ is the sequence length, and $C$ is the feature dimension at each nucleotide position, is processed by a two-layer LSTM. The final hidden state is then used as the LSTM representation.

For feature fusion, the CNN and LSTM outputs are concatenated and passed through an attention mechanism. The attention weights are computed using a two-layer network with a tanh activation, and used to weight the two branches:

\begin{equation}
F_{\text{fused}} = \alpha \cdot F_{\text{cnn}} + (1 - \alpha) \cdot F_{\text{lstm}}
\end{equation}
Subsequently, a linear transformation and LayerNorm are used to further unify the dimensions, which is taken as the sequence embedding representation:
\begin{equation}
F_{\text{seq}} = \operatorname{LayerNorm}(\operatorname{ReLU}(W_f \cdot F_{\text{fused}}))
\end{equation}

\subsection{HyperGraphMamba Multimodal Feature Fusion}

We propose HyperGraphMamba for multimodal feature fusion. First, we leverage Mamba’s linear-complexity design~\cite{gu2023mamba} to efficiently align and enhance intra-modal representations. Then, each modality is treated as a node, and virtual hypernodes are introduced to form a multimodal hypergraph, enabling rich cross-modal interaction. To our knowledge, this is the first use of hypernodes in multimodal fusion. The framework consists of two parts: Mamba-based intra-modal modeling and hypergraph-based inter-modal interaction.They are introduced as follows:

\subsubsection{Mamba Intra-modal Modeling}
Firstly, embedding $F_m$ are extracted from the sequence, structure, and expression modalities respectively, where m $\in {(\text{seq}, \text{str}, \text{exp}) }$.

After linear projection, each modality is refined by a lightweight Mamba module. Beyond sequence modeling, Mamba unifies feature scales, enriches intra-modal representations, and provides high-quality inputs for effective hypergraph fusion. This is achieved via two independent state space sub-networks:

\begin{equation}
H^{\text{intra}}_m = \operatorname{SSM}^{\rightarrow}_m(Z_m) + \operatorname{flip} \left( \operatorname{SSM}^{\leftarrow}_m \left( \operatorname{flip}(Z_m) \right) \right)
\end{equation}
where $\operatorname{SSM}^{\rightarrow}_m(\cdot)$ and $\operatorname{SSM}^{\leftarrow}_m(\cdot)$ respectively represent the left-to-right and right-to-left state space modeling modules for modality , using structural variants of Mamba; $\operatorname{flip}(\cdot)$ denotes reversing the order of tokens in the sequence.

\subsubsection{Hypergraph Fusion Mechanism}

In the inter-modal interaction stage, a multimodal hypergraph is built from the intra-modal representations $H^{\text{intra}}_{\text{seq}}$, $H^{\text{intra}}_{\text{str}}$, and $H^{\text{intra}}_{\text{exp}}$, serving as initial node features. The hypergraph $\mathcal{H} = (\mathcal{V}, \mathcal{E}_H)$ includes three modality nodes and $K$ learnable virtual nodes to enhance high-order structural modeling. Edge weights are based on cosine similarity, and the Laplacian matrix $L$ is computed with normalization.
\begin{equation}
L_{ij} = \frac{\cos(v_i, v_j)}{\sqrt{\sum_k \cos(v_i, v_k)} \cdot \sqrt{\sum_k \cos(v_j, v_k)}}
\end{equation}

The node features are then processed by a residual GCN with hierarchical attention to produce updated modality-specific embeddings $Z^{\text{hyper}}_m$.

Next, we extract the [CLS] token and mean pooling vector from each modality, concatenate them, and apply a linear projection to obtain modality embeddings:

\begin{equation}
F_m = \operatorname{Linear}\left( \left[ Z^{\text{hyper}}_m[\text{CLS}] \,\|\, \operatorname{MeanPool}(Z^{\text{hyper}}_m) \right] \right)
\end{equation}

Finally, the fused representation is formed by concatenating all modalities:
$F_{\text{fused}} = [F_{\text{seq}} \,\|\, F_{\text{str}} \,\|\, F_{\text{exp}}]
$. This module supports two-modal or three-modal interaction and can adapt to scenarios with missing modalities.
 
\subsection{ncRNA Classification}

We design a shallow feedforward network for ncRNA multi-class classification. The output of the $l$-th fully connected layer is defined as $H^{(l)}$. 

\begin{equation}
H^{(l)} = \operatorname{ReLU} \left( \operatorname{BN}\left(W^{(l)} H^{(l-1)} + b^{(l)}\right) \right)
\end{equation}
where $H^{(0)} = F_{\text{fused}}$. A final linear layer maps $H^{(L)}$ to class logits:
\begin{equation}
Y_{\text{logits}} = W^{(\text{out})} \cdot H^{(L)} + b^{(\text{out})}
\end{equation}

Softmax is applied to obtain class probabilities $\hat{y}$, and the model is optimized using cross-entropy loss:

\begin{equation}
\mathcal{L}_{\text{CE}} = -\frac{1}{B} \sum_{i=1}^{B} \sum_{c=1}^{C} y_{i,c} \log \hat{y}_{i,c}
\end{equation}

\begin{table}[htbp]
\centering
\scriptsize
\begin{tabular}{>{\centering\arraybackslash}m{0.4cm}lccccc}
\toprule
Dataset & Method & ACC & MCC & F1-Score & Recall & Precision \\
\midrule
\multirow{7}{*}{D1} 
& MMnc         & \underline{0.9880} & \underline{0.9860} & \underline{0.9850} & \underline{0.9880} & \underline{0.985} \\
& BioDeepFuse  & 0.7634 & 0.7187 & 0.7374 & 0.7634 & 0.7600 \\
& MFPred       & 0.9632 & 0.9561 & 0.9506 & 0.9527 & 0.9552 \\
& ncrna-deep   & 0.9725 & 0.9668 & 0.9670 & 0.9680 & 0.9662 \\
& RNAGCN       & 0.9061 & 0.8956 & 0.9056 & 0.9052 & 0.9073 \\
& RiNALMo      & 0.8564 & 0.8267 & 0.8565 & 0.8565 & 0.8531 \\
& HGMamba-ncRNA          & \textbf{0.9920} & \textbf{0.9910} & \textbf{0.9910} & \textbf{0.9910} & \textbf{0.9910} \\
\midrule
\multirow{7}{*}{D2}
& MMnc         & \underline{0.9650} & 0.9410 & 0.9320 & 0.9210 & 0.9550 \\
& BioDeepFuse  & 0.8106 & 0.7887 & 0.8059 & 0.8106 & 0.8033 \\
& MFPred       & 0.9590 & \underline{0.9544} & \underline{0.9644} & \underline{0.9628} & \underline{0.9670} \\
& ncrna-deep   & 0.9091 & 0.8992 & 0.8995 & 0.8979 & 0.9101 \\
& RNAGCN       & 0.9157 & 0.9062 & 0.9088 & 0.9091 & 0.9096 \\
& RiNALMo      & 0.5751 & 0.5264 & 0.5625 & 0.5751 & 0.5682 \\
& HGMamba-ncRNA          & \textbf{0.9720} & \textbf{0.9690} & \textbf{0.9680} & \textbf{0.9690} & \textbf{0.9770} \\
\midrule
\multirow{7}{*}{D3}
& MMnc         & 0.9670 & 0.9440 & 0.9300 & \underline{0.9520} & 0.9190 \\
& BioDeepFuse  & 0.8845 & 0.8042 & 0.8646 & 0.8845 & 0.8790 \\
& MFPred       & \underline{0.9705} & \underline{0.9493} & \underline{0.9350} & 0.9249 & \underline{0.9480} \\
& ncrna-deep   & 0.9562 & 0.9242 & 0.9052 & 0.8935 & 0.9184 \\
& RNAGCN       & 0.9283 & 0.8775 & 0.8043 & 0.7880 & 0.8375 \\
& RiNALMo      & 0.8895 & 0.8112 & 0.8774 & 0.8895 & 0.8754 \\
& HGMamba-ncRNA          & \textbf{0.9780} & \textbf{0.9630} & \textbf{0.9540} & \textbf{0.9460} & \textbf{0.9670} \\
\bottomrule
\end{tabular}
\caption{Performance comparison of different methods on D1, D2, and D3 datasets.}
\label{tab:multiset_comparison}
\end{table}
\section{Experiments And Results}
\subsection{Datasets}
We evaluate model performance on three datasets: D1, D2, and D3.
D1, from Ruiting Xu et al., includes RNA sequences and secondary structures, with 29,306 non-coding RNA entries from mammals (mainly mice), spanning 7 classes.
D2, collected by Lima et al., also contains sequences and structures, comprising 45,447 entries across 13 classes.
D3, provided by Constance et al., includes sequence, structure, and expression features, with 19,131 samples from cancerous and healthy tissues, covering 4 types of non-coding RNAs. 

\subsection{Comparison Baselines}

To evaluate the effectiveness of HGMamba-ncRNA, we compare it with six representative baselines (MMnc, BioDeepFuse, MFPred, ncrna-deep, RNAGCN, and RiNALMo) on three datasets. As shown in Table~\ref{tab:multiset_comparison}, HGMamba-ncRNA achieves the best performance across all datasets and metrics, demonstrating superior classification and generalization ability.


On the single-species dataset D1, while most models perform well, our method still leads, reflecting strong fundamental feature modeling. For D2, with a larger and more diverse dataset, methods like BioDeepFuse and RiNALMo degrade significantly, indicating limited generalization. In contrast, our model maintains stable performance, showcasing robustness under complex distributions. D3 includes expression data and greater structural heterogeneity. Only MMnc and our method utilize all three modalities. By better integrating expression features, our model outperforms MMnc, underscoring its strength in multimodal fusion.

\subsection{Hyperparameters}
To assess the robustness and sensitivity of HGMamba-ncRNA, we conduct hyperparameter experiments on three core modules—MKCNN, CPKAN, and MultiTransformer—using the D3 dataset. Main hyperparameter settings are summarized and detailed analysis is in Appendix A. The results are presented in Figure~\ref{fig:3}.

\subsubsection{MKCNN Module}

The MKCNN module is designed to capture both local sequence features (e.g., stem-loops, binding sites) and long-range dependencies in RNA sequences. To achieve a balance between resolution and abstraction, we evaluated several multi-scale CNN configurations. A 4-layer CNN with kernel sizes [3,5,7,9] and channel dimensions [128,256,512,1024] was selected due to its stable and robust performance. For modeling global contextual dependencies, we explored LSTM depths of \{2, 3, 4, 5\}. A 3-layer LSTM was ultimately adopted, providing an effective compromise between modeling capacity and training stability.

\subsubsection{CPKAN Module}



The CPKAN module models the RNA expression modality, which is high-dimensional, noisy, and shaped by gene co-regulation. We varied the Chebyshev polynomial degree in \{1, 2, 3, 4, 5, 6\} and found degree 5 best balances expressiveness and noise suppression. For scaling, scale\_base and scale\_cheby were tested in \{0.5, 1, 2, 5\}, with 1 chosen for both to enhance nonlinearity while preserving stability and avoiding overfitting.

\subsubsection{MultiTransformer Module}

The MultiTransformer module captures RNA structural features, including long-range base pairings and multi-scale topological dependencies. We tested attention heads in \{1, 2, 4, 8\}, selecting 4 to balance diversity and stability. For model depth, Transformer layers \{1, 2, 3, 4, 5\} were evaluated, with 3 layers chosen to ensure sufficient structural integration without over-smoothing. To model hierarchical base-pairing, we evaluated scale settings \{[1], [1,2], [1,2,3]\}, with [1,2,3] providing effective multi-level representation while avoiding noise.

\subsection{Ablation Study}
To evaluate the contribution of each submodule to the overall model performance, we conducted a comprehensive ablation study. This includes evaluating the architectural design of the MKCNN-LSTM sequence encoder, alternative encoding strategies for expression features in the CPKAN module, and the multi-scale fusion mechanism in the Multi-Scale Transformer for structural feature extraction.
\begin{figure*}
\centering
\includegraphics[width=\textwidth]{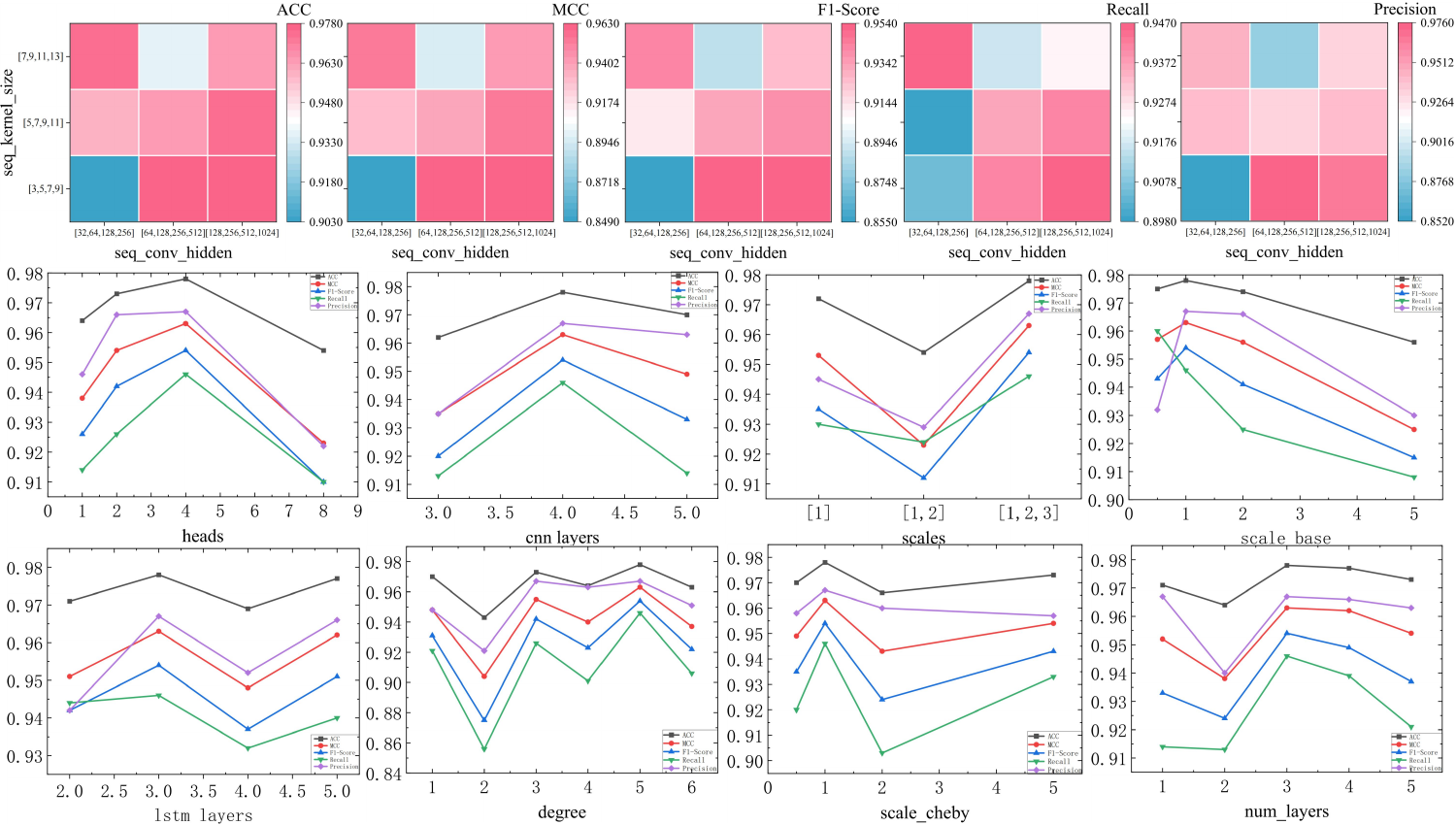}
\caption{Results of hyperparameter analysis.
}
\label{fig:3}
\end{figure*}
\subsubsection{Modality Contribution Analysis}

To assess the contribution of each modality, we evaluated classification performance under four modality combinations: sequence+structure (Figure 4(A.1)), sequence+expression (Figure 4(A.2)), structure+expression (Figure 4(A.3)), and all three modalities (Figure 4(A.4)).

The tri-modal configuration yields the highest overall accuracy, showing that integrating all modalities enhances robustness and reduces the impact of missing information.
Notably, the structure+expression setting—lacking sequence data—achieves the lowest overall accuracy but excels in miRNA classification (0.87 vs. 0.25 for snoRNA), suggesting that expression features are particularly informative for miRNA-specific patterns. In comparison, full-modal fusion yields notable improvements, especially for challenging classes like snoRNA (accuracy improves from 0.82 to 0.83). lncRNA and snRNA also achieve near-perfect accuracy (0.99 and 0.98), underscoring the critical role of sequence data, while structural and expression features complement functional characterization.

In summary, the three modalities offer complementary strengths—sequence provides nucleotide-level context, structure encodes spatial configurations, and expression captures dynamic biological states—jointly enhancing classification accuracy and model robustness.
\subsubsection{Intra-Modality Method Analysis}
\paragraph{1) Sequence Feature Extraction.}

\begin{figure*}[t!]
\centering
\includegraphics[width=\textwidth]{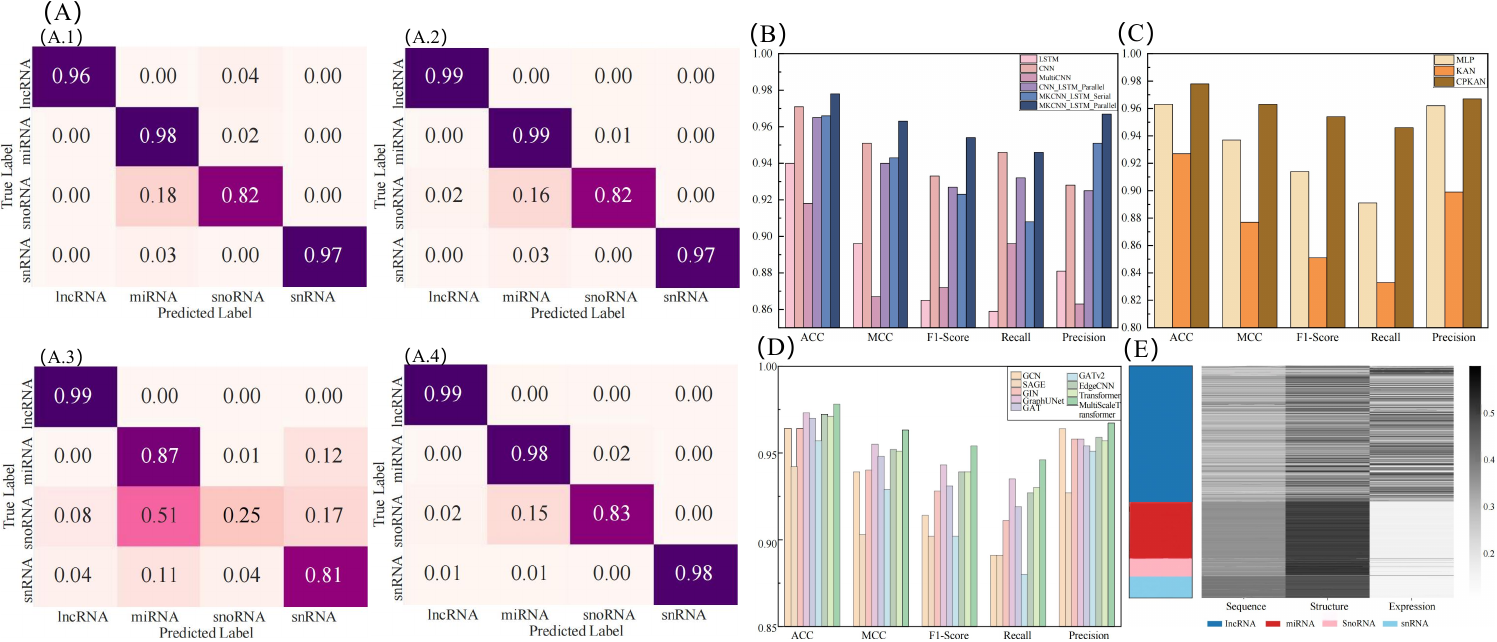}
\caption{Ablation experiment results for ablation study.
}
\label{fig:5}
\end{figure*}

We compare several architectures for sequence modeling, including LSTM, CNN, multi-kernel CNN (MKCNN), parallel CNN+LSTM, and sequential MKCNN-LSTM (Figure~\ref{fig:5}(B)). LSTM alone underperforms due to weak local pattern capture; CNN improves local extraction but lacks contextual modeling. MKCNN enriches multi-scale features but suffers from absent fusion. Parallel CNN+LSTM improves performance but increases complexity, while MKCNN-LSTM sequentially incorporates context but loses temporal details due to static MKCNN output. In contrast, our method effectively integrates local and global semantics, yielding superior and more generalizable performance.

\begin{figure}[t!]
\centering
\includegraphics[width=0.45\textwidth]{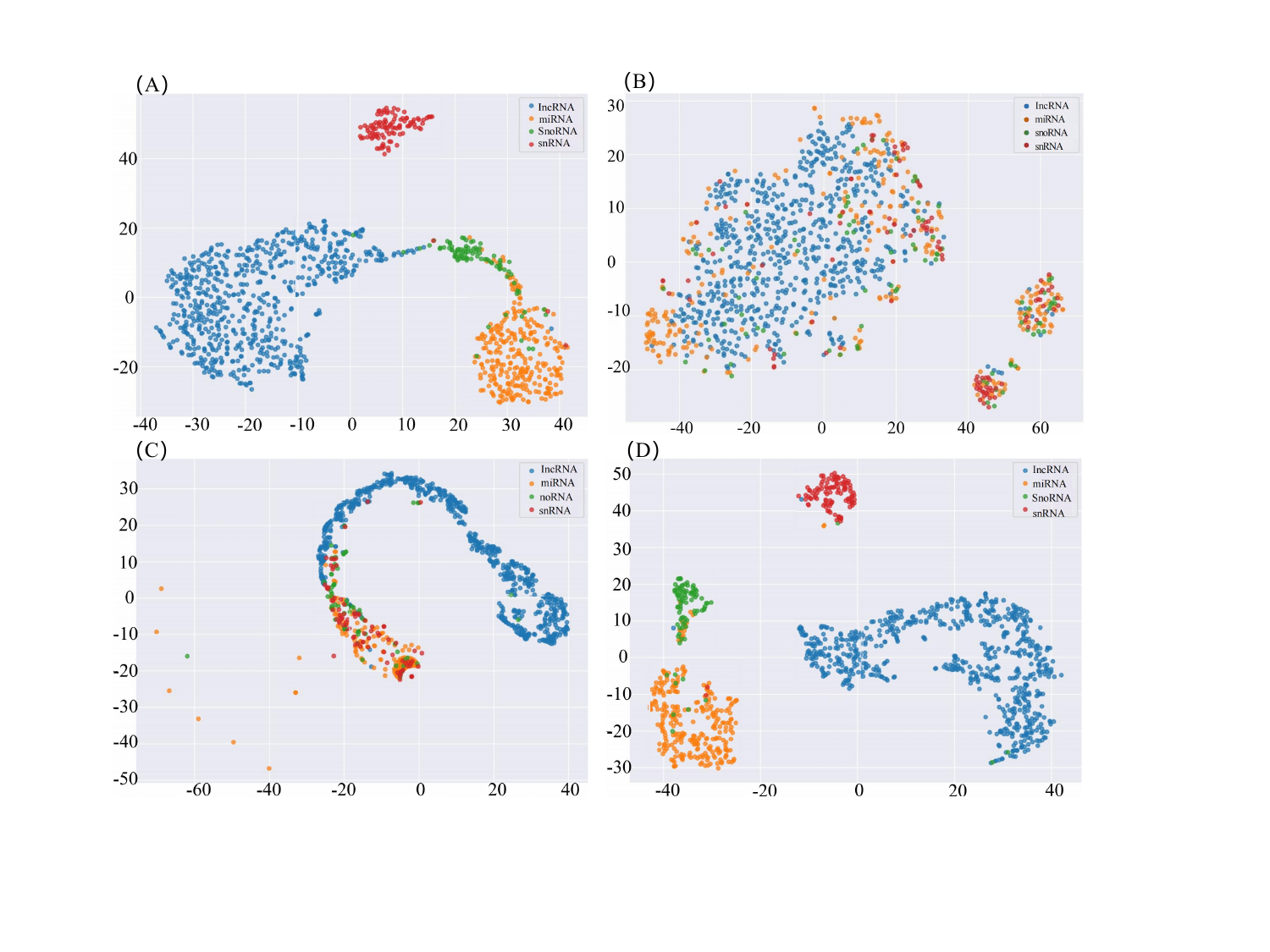}
\caption{Visualize of the feature space distributions of (A) sequence, (B) structure, (C) expression, and (D) fused features, respectively.
}
\label{fig:10}
\end{figure}
\paragraph{2) Expression Feature Extraction}

For expression data modeling, we compare our method with two representative baselines: classical Multilayer Perceptron (MLP) and KAN. Rresults are shown in Figure~\ref{fig:5}(C).
CPKAN outperforms both across all metrics, especially in Precision, MCC, and F1-score. RNA expression data are high-dimensional, noisy, and exhibit nonlinear, frequency-domain characteristics. MLP offers basic nonlinear modeling but fails to capture localized and frequency-specific patterns. KAN lacks adaptability due to fixed kernels, yielding the weakest results. In contrast, CPKAN uses Chebyshev polynomials to model fine-grained, oscillatory variations, enhancing its ability to detect both subtle and abrupt regulatory signals.

\paragraph{3) Structural Feature Extraction}


To assess the impact of structural modeling on RNA classification, we compare our structural module with representative GNNs: GCN, GAT, GATv2, GraphSAGE, GIN, GraphUNet, EdgeCNN, and Graph Transformer (Figure~\ref{fig:5}(D)).
GCN and GIN rely on fixed-order neighborhood aggregation, limiting their ability to model long-range base pairings in RNA structures. GraphSAGE improves flexibility but still focuses on local features. GAT and GATv2 introduce attention mechanisms, partially alleviating this limitation, but struggle under single-scale settings. GraphUNet’s pooling captures hierarchical features but may discard critical information in small RNA graphs. EdgeCNN effectively models edge features but lacks adaptability to heterogeneous graph structures. Graph Transformer improves long-range dependency modeling via global attention. Building on this, our MultiScale Graph Transformer introduces multi-hop neighborhood aggregation and hierarchical stacking, enabling joint modeling of local and global RNA topologies. This significantly enhances structural representation capacity and outperforms all baseline methods.
\subsubsection{Fusion Module Analysis}
To explore modality preferences during fusion, we visualize attention weights across sequence, structure, and expression on the D3 dataset (Figure~\ref{fig:5}(E)). Each heatmap row represents an ncRNA sample, with darker colors indicating higher attention.

Overall, the structure modality consistently receives higher attention across classes, highlighting its strong discriminative power—especially for snRNA and snoRNA. lncRNAs rely on both structure and expression, reflecting structural interfaces and expression specificity. miRNAs emphasize sequence and structure, linked to conserved motifs and folding. snoRNAs focus mainly on structure, consistent with their stable forms (e.g., C/D-box, H/ACA-box). snRNAs attend to both structure and sequence, aligning with their conserved features.

\subsection{Cross-Dataset Generalization}

To assess cross-dataset generalization, we fix D1 (mammals) as the test set and use D2 and D3 as heterogeneous training sets. Experimental details and analysis are in the Appendix B. Results show that our multimodal model generalizes well across datasets, especially when structural consistency enables effective transfer of RNA representations.

\subsection{Visualization Analysis}


To better understand the discriminative capability of each modality, we performed t-SNE visualization on the extracted features from sequence, structure, and expression modalities, as well as their fused representation. As shown in Figure~\ref{fig:10}, the sequence modality separates miRNA and snRNA well but shows partial overlap between lncRNA and snoRNA, limiting its overall discriminative power. The structure modality reveals clusters for miRNA and snRNA, yet suffers from considerable mixing, especially for lncRNA samples. The expression modality displays the weakest separability, with substantial overlap across all classes. In contrast, the fused feature space exhibits clearly separated clusters for all four RNA types, highlighting the effectiveness of multimodal fusion in enhancing representation learning.
\section{Conclusion}
In this study, we propose HGMamba-ncRNA, a novel model for functional classification of non-coding RNAs (ncRNAs), integrating sequence, secondary structure, and expression features. It combines key modules—MKC-L, MSGraphTransformer, CPKAN, and HyperGraphMamba—to effectively extract and fuse multimodal information. Experiments on benchmark datasets show that our method outperforms state-of-the-art approaches in accuracy, robustness, and generalization. Ablation studies and feature visualization further validate the effectiveness of our fusion strategy, offering a promising framework for automated ncRNA identification and analysis.



\bibliography{aaai2026}

\end{document}